\documentclass[runningheads]{llncs}
\usepackage{graphicx}
\usepackage{amsmath}
\usepackage{amssymb}
\usepackage[ruled,vlined]{algorithm2e}
\usepackage{algorithmic}
\usepackage{bbm}
\usepackage{booktabs}
\usepackage{multirow}
\usepackage{soul,xcolor}
\usepackage{subfig}

\begin{document}
\title{BAMAX: Backtrack Assisted Multi-Agent Exploration using Reinforcement Learning}

\titlerunning{BAMAX}

\author{Geetansh Kalra\inst{1} \and
Amit Patel\inst{1} \and
Atul Chaudhari\inst{1} \and
Divye Singh\inst{1}}

%
%
\institute{Engineering for Research
Thoughtworks India Pvt Ltd.
Pune, India}

\maketitle              
%

\begin{abstract}

Autonomous robots collaboratively exploring an unknown environment is still an open problem. The problem has its roots in coordination among non-stationary agents, each with only a partial view of information. The problem is compounded when the multiple robots must completely explore the environment. In this paper, we introduce Backtrack Assisted Multi-Agent Exploration using Reinforcement Learning (BAMAX), a method for collaborative exploration in multi-agent systems which attempts to explore an entire virtual environment. As in the name, BAMAX leverages backtrack assistance to enhance the performance of agents in exploration tasks. To evaluate BAMAX against traditional approaches, we present the results of experiments conducted across multiple hexagonal shaped grids sizes, ranging from 10x10 to 60x60. The results demonstrate that BAMAX outperforms other methods in terms of faster coverage and less backtracking across these environments.

\end{abstract}

\section{Introduction}
\label{sec:introduction}

Autonomous exploration by robots in unknown environments has  diverse applications such as search and rescue, environmental monitoring, and disaster management, and is still an open challenge \cite{Liu2007-ir}. Moreover, individual robots often struggle with limitations in coverage, efficiency, reliability, resiliency, and adaptability when operating in complex and dynamic environments.

To overcome these challenges, multi-agent collaborative systems have gained attention \cite{doi:10.5772/57313}. By leveraging collective knowledge and coordinating actions, these multiple agents can explore the environment more effectively, leading to improved coverage, robustness, and information exchange \cite{tan1993multi,10.1007/978-3-319-71682-4_5}. However, collaborative strategies may encounter challenges such as navigating local extrema or overcoming dead ends \cite{Schultz2002MultiRobotSF}.

In this paper we present an approach called \textit{Backtrack Assisted Multi-Agent Exploration using Reinforcement Learning} (BAMAX, for short). Multi-agent reinforcement learning is chosen as the foundation for our approach due to its capability to enable agents to learn optimal behaviors through interactions with the environment, leveraging rewards and penalties to enhance decision-making abilities. Our method allows multiple robots to autonomously explore and construct maps within hexagonal mazes, incorporating the crucial capability to backtrack to previously known open positions when encountering obstacles.

This paper has two main contributions as follows. The first one is the guarantee of full exploration. Our approach leverages the collective abilities of multiple robots to facilitate efficient navigation, overcome walls, and achieve complete coverage of the entire grid. The second contribution is an ability to scale to multiple sizes of hexagonal grids. In the rest of the paper, we will discuss how these contributions are made using the BAMAX method.


The section \ref{sec:related-work} provides an overview of the related works, covering both traditional methods and the application of intelligent agent navigation with Deep Reinforcement Learning (DRL). In section \ref{sec:bamax}, we present our approach, detailing the creation of the environment that addresses our specific problem statement, and the setup of the Reinforcement Learning (RL) framework. Additionally, this section delves into the specifics of our proposed algorithm and the underlying network architecture. The section \ref{sec:experiments-and-results} focuses on the experimentation conducted to evaluate the performance of our method in comparison to other traditional approaches. Thereafter, the section discusses the results obtained from our experiments, offering analysis and comparison of our method with alternative approaches. Finally, the section \ref{sec:conclusion} concludes the paper.


\section{Related Work}
\label{sec:related-work}


The traditional methods for autonomous exploration in unknown environments identify frontier cells as boundaries between known and unknown areas. In an approach discussed in the work \cite{613851}, each robot maintains a global evidence grid, integrating its local grid with the global map at the frontiers. By summing log odds probabilities, a collaborative and decentralized system is achieved. Another approach \cite{zlot2002multi} utilizes a bidding protocol for assigning regions with high unknown areas to robots. The work discussed in \cite{González} proposed a navigation approach with safety regions and a \textit{Next-Best-View} algorithm.

While traditional methods have shown success, incorporating high-level knowledge like structural patterns remains a challenge \cite{9359174}. To address this challenge and to improve exploration strategies in unknown environments, recent advancements in \textit{Deep Reinforcement Learning} (DRL, for short) offer potential. A novel approach introduced in the work \cite{tai2016cognitive} leverages raw sensor data from an RGB-D sensor to develop a cognitive exploration strategy through end-to-end DRL. Similarly, the D3QN algorithm discussed in the work \cite{8832393} enables mobile robots to gradually learn about their environment and autonomously navigate to target destinations using only a camera, avoiding obstacles. 

In dynamic environments, \cite{long2018optimally} employed a deep neural network with long short-term memory and a reinforcement learning algorithm for effective robot navigation. Another study by \cite{tai2017virtualtoreal} utilized environment information as input to a neural network and trained agents using the asynchronous deep deterministic policy gradient algorithm. To extract and utilize structural patterns, \cite{8463213} introduced convolutional networks to encode such information and employed the A3C algorithm during training. 
\section{Our Approach}
\label{sec:bamax}

In this section, we will discuss the hexagonal grid environment used in our experimentation and the setup of reinforcement learning for our problem. We will also explain the architecture and how data flows in our method, BAMAX. 

\subsection{Environment}
\label{sec:bamax:environment}

\begin{figure}[h]
    \centering
    \includegraphics[width=0.4\textwidth,height=0.5\textwidth]{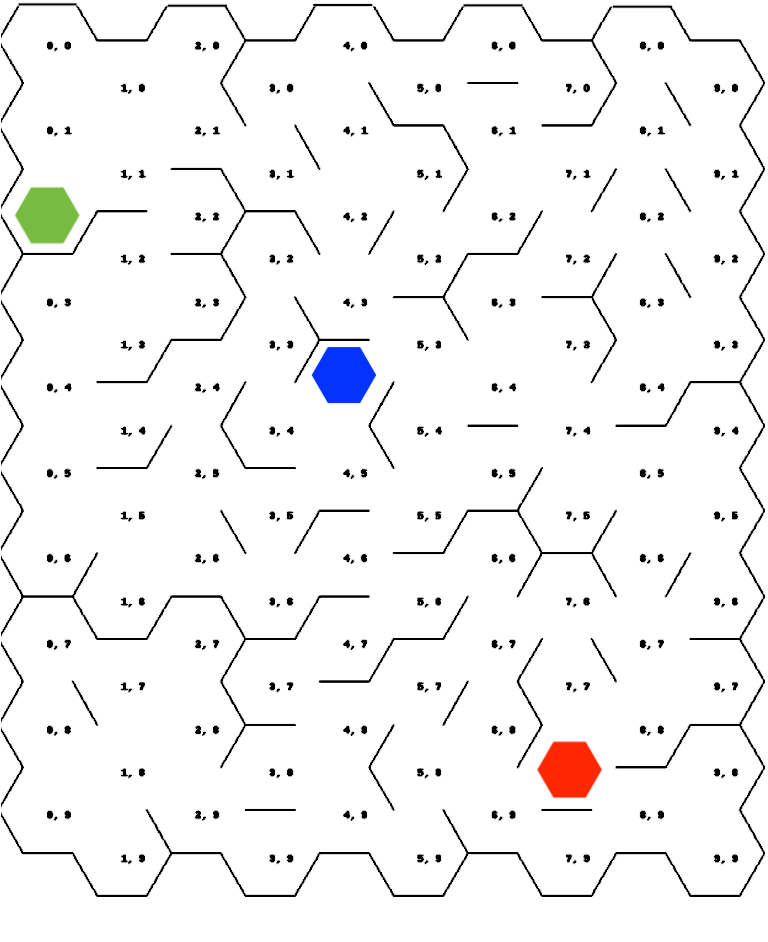}
    \caption{Hexagonal environment of size $\mathcal{G}_{10}$ with three agents represented by green, blue, and red hexagons}
    \label{fig:hex-env-with-agents}
\end{figure}

The environment used in this paper is a two-dimensional grid maze, $\mathcal{G}_d$, with \textit{d} number of hexagonal cells across height and width. The hexagonal shape was selected for the unit cell in our environment because out of all regular polygons, hexagon is the shape with the highest number of edges which creates a regular tiling in the euclidean plane (other being square and triangle). This would facilitate consistency while also providing a higher complexity in terms of choice of directions available to move from one cell to another. Moreover, to allow for distinct ingress and egress points for each cell, we ensure that more than two sides are left open while creating the maze. An illustration of our hexagonal grid maze ($\mathcal{G}_{10}$) can be seen in figure \ref{fig:hex-env-with-agents}.


\subsection{Building A Reinforcement Learning Model}
\label{sec:bamax:model}

In our study, we utilize the DQN algorithm \cite{mnih2013playing} to approximate the Q-function, which represents the optimal action-value function using a deep neural network architecture. By taking the agent's observations as input, the DQN architecture generates Q-values for various actions. These Q-values indicate the expected cumulative rewards that the agent can attain by taking specific actions in a given state. The agent's objective is to select actions with higher Q-values, thereby maximizing long-term rewards and making optimal decisions in the environment (Eq \ref{eq: Q-value}). 


\begin{equation}
Q(s, a) = \mathbb{E} \left[ r + \gamma \max_{a'} Q(s', a') \,|\, s, a \right]
\label{eq: Q-value}
\end{equation}

\subsubsection{State Space} 
To capture the information contained in the environment effectively, we extracted the following six distinct sub-states.

\begin{enumerate}
    \item $s_{\text{exp}}$: Image representing the area of the map that has been explored.
    \item $s_{\text{unexp}}$: Image corresponds to the area of the map that is yet to be explored.
    \item $s_{\text{agents-pos}}$: Image indicating the location of the agent in the map.
    \item $s_{\text{other-agents-pos}}$: Image showing the positions of other agents in the map.
    \item $s_{\text{walls}}$: Image containing the information about the walls present in the explored area.
    \item $s_{\text{local-obs}}$: This component provides a local observation for each agent, capturing essential information such as the presence of walls, the presence of predators, and whether neighboring cells have already been explored in each direction relative to the agent's current position.
\end{enumerate}
This decomposition allows us to utilize relevant features from different aspects of the environment to enhance the learning and decision-making capabilities of our method. 
These sub-states together form a single state and the entire state space, $\mathcal{S}$ for our method is represented as \ref{eq:state}.

\begin{equation}
\begin{aligned}
\text{State space, }\mathcal{S} = \{s^k = (s_{\text{exp-map}}, s_{\text{unexp-map}}, s_{\text{agents-own-pos}}, \\
s_{\text{other-agents-pos}}, s_{\text{walls-map}}, s_{\text{local-obs}}) \; | \; k \in \{agents\}\}
\end{aligned}
\label{eq:state}
\end{equation}

\subsubsection{Action Space}


The action space \(A\) consists of six distinct actions, denoted by \(a_i\) available to the agent. The actions correspond to the movement of the agent along the six sides of a hexagon cell. 

\subsubsection{Reward}
In our specific problem, we aimed our agent to consider increasing rewards through exploration and incentivize moving toward unexplored regions to maximize overall exploration. In order to capture this, we formulated the reward as comprising of two components as seen in equation \ref{eq: total-reward}

\begin{equation}
r_t = r_{\text{immediate},t} + r_{\text{surrounding},t}
\label{eq: total-reward}
\end{equation}

$r_{\text{immediate}, t}$ is given immediately after the agent takes an action. If the agent enters a valid cell, it receives a reward based on the size of the explored area and the number of unexplored cells. It encourages exploration by providing higher rewards for more unexplored cells. If the agent bumps into another predator or a wall, it receives a penalty.

\begin{equation}
r_{\text{immediate}, t} = \begin{cases}
\max\left(50, \text{explored\_cells - unexplored\_cells}\right) & \text{valid cell} \\
-20 & \text{collision}
\end{cases}
\end{equation}


$r_{\text{surrounding}, t}$ is calculated based on the number of immediate unexplored cells visible to the agent. It considers the agent's local observation and promotes exploration by providing rewards for unexplored cells within the agent's visibility range.

\begin{equation}
r_{\text{surrounding}, t} = \sum_{x \in \text{neighbouring-cell}} \mathbb{I}( x = \text{unexplored\_cell} )
\end{equation}

These reward components aim to guide the agent towards unexplored areas, incentivizing exploration and the acquisition of new information.

\subsection{Backtrack Assisted Multi-Agent Exploration using Reinforcement Learning (BAMAX)}

\begin{figure}[!hb]
    \centering
    \includegraphics[width=1\textwidth]{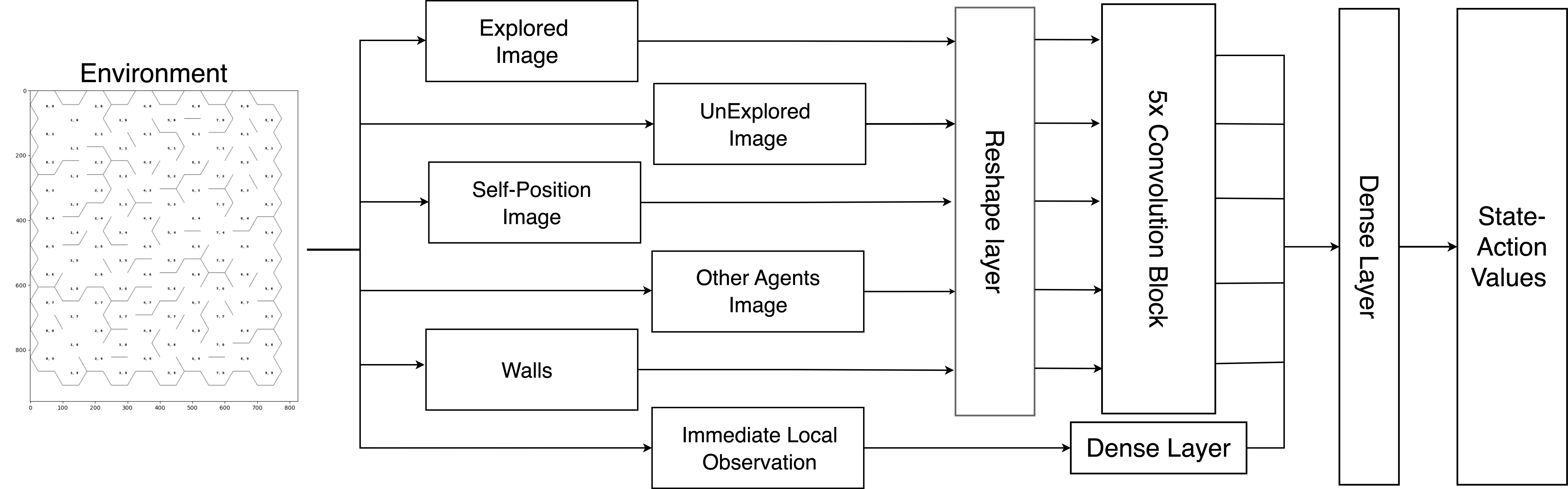}
    \caption{Architecture of BAMAX}
    \label{fig: arch-of-BAMAX}
\end{figure}

In this section, we discuss the architecture of BAMAX as shown in the figure \ref{fig: arch-of-BAMAX}.
In our approach, we decompose the environment into six parts to enhance representation, as described in section \ref{sec:bamax:model}.

\textbf{Image Reshape Layer} receives the five of these six parts as the input, and transforms the state space images of varying sizes into a standardized fixed size. By incorporating the Image Reshape Layer, our architecture maintains flexibility and adaptability to different environment sizes.

\textbf{CNN Layer} utilizes \textit{Convolutional Neural Networks} (CNNs) \cite{NIPS2012_c399862d} to extract meaningful features from the environment observations in BAMAX. Each image input undergoes a specific sequence of convolutional layers, activation functions, and pooling layers, as depicted in the table \ref{tab:processing-steps}.

\begin{table}[h]
\caption{Processing Steps for Different Image Inputs}
\setlength{\tabcolsep}{8pt} 
\label{tab:processing-steps}
\begin{tabular}{@{}lllll@{}}
\toprule
\textbf{Image Type} &
  \textbf{\begin{tabular}[c]{@{}l@{}}Layer \\ Type\end{tabular}} &
  \textbf{\begin{tabular}[c]{@{}l@{}}Kernel \\ Size\end{tabular}} &
  \textbf{\begin{tabular}[c]{@{}l@{}}Kernel \\ No.\end{tabular}} &
  \textbf{Activation} \\ \midrule
\multirow{2}{*}{Explored Image}              & Convolution & 1x1 & 32 & ReLu \\ \cmidrule(l){2-5} 
                                             & Max Pooling & 2x2 &    &  \\ \midrule
\multirow{2}{*}{Unexplored Image}            & Convolution & 1x1 & 32 & ReLu \\ \cmidrule(l){2-5} 
                                             & Max Pooling & 2x2 &    &  \\ \midrule
\multirow{2}{*}{Agents Own Position Image}   & Convolution & 1x1 & 32 & ReLu \\ \cmidrule(l){2-5} 
                                             & Max Pooling & 2x2 &    &  \\ \midrule
\multirow{2}{*}{Other Agents Position Image} & Convolution & 1x1 & 32 & ReLu \\ \cmidrule(l){2-5} 
                                             & Max Pooling & 2x2 &    &  \\ \midrule
\multirow{4}{*}{Walls Image}                 & Convolution & 5x5 & 32 & ReLu \\ \cmidrule(l){2-5} 
                                             & Max Pooling & 4x4 &    &  \\ \cmidrule(l){2-5} 
                                             & Convolution & 5x5 & 64 & ReLu \\ \cmidrule(l){2-5} 
                                             & Max Pooling & 4x4 &    &  \\ \midrule
Flatten                                      &             &     &    &      \\ \bottomrule
\end{tabular}
\end{table}

\textbf{Local Observation Layer} takes in the local observation of the agent and passes it through two of the dense layers with 62 and 32 neurons respectively, with a ReLU activation applied to each of them 

\textbf{Aggregation Layer and Action Layer} takes in the output of the local observation layer and combines it with the flattened outputs of all the environment observations from the CNN layer component. This combined representation is fed into another dense layer with 32 neurons and a ReLU activation. Finally, the output of this layer is passed through a dense layer with a size equal to the desired action space, using a linear activation function. This final layer produces the Q-values associated with each action, allowing the agent to choose the most advantageous action based on the learned policy. 


\textbf{Backtrack Assistance} mechanism addresses situations where agents become trapped or stuck within already explored regions. Its primary purpose is to facilitate the agents' navigation back to the last unvisited node or cell, providing them with an escape route from local extrema or dead ends. As the agents explore the environment, they collectively store the connections between cells using a graph data structure, where each cell is represented as a node and the connections between cells are represented as edges. This representation enables the entire environment to be visualized as a graph, which in turn allows the application of the A* algorithm \cite{4082128} to guide the agents in backtracking to the nearest unvisited node. By backtracking to unexplored areas, the agents can resume the exploration process and continue discovering new information, enabling them to expand their understanding of the environment and potentially uncover valuable insights.

\textbf{Training Strategy} employs a centralized training approach by utilizing a single network to train all agents. A single replay buffer stores the experiences of all agents and the training samples are sampled from this shared replay buffer during the training process. Further, $\epsilon$-greedy approach is used to maintain the balance between exploration and exploitation. The whole training process is also depicted in algorithm \ref{Algo: BAMAX Algo}

\begin{algorithm}[t]
    \SetKwInOut{Input}{Input}
    \SetKwInOut{Output}{Output}
    
    \Input{Episode number, Flag indicating if all agents are alive}
    \Output{None}
    
    \BlankLine
    Initialize replay buffer $B$ with size $|B|$\;
    Initialize BAMAX\;
    
    \For{episode in range(Total Episodes)}{
        Reset the grid and create agents\;
        Create an empty graph $G$\;
        
        Agents receive initial observation states $s_t$\;
        
        \For{timestep in range(timesteps per episode)}{
            \If{Flag indicating if all agents are alive}{
                \For{agent $u \in U$}{
                    \If{agent is stuck}{
                        Find nearest unexplored node and guide agent using A*\;
                    }
                    \Else{
                        Get agent's observation $o_{t}^u$\;
                        Get action $a_{t}^u$\;
                        Update $s_{t+1}^u$\;
                        Update $o_{t+1}^u$, $r_{t}^u$\;
                        Store transition $(s_{t}^u, a_{t}^u, r_{t}^u, s_{t+1}^u)$ in $B$\;
                    }
                }
                Check if all cells are explored or any agent is terminated\;
                
                \If{termination condition is met}{
                    Set the flag to false\;
                    
                    \If{episode number $\geq$ 50}{
                        Sample a batch of experiences from $B$\;
                        Update Q value using $Q(s_i, a_i) = r_i + \gamma \cdot \max_{a'}(Q(s_{i+1}, a'))$
                    }
                }
            }
            \Else{
                Break out of the loop\;
            }
        } \textbf{end}
    } \textbf{end}
    
    \caption{BAMAX Algorithm}
    \label{Algo: BAMAX Algo}
\end{algorithm}

\textbf{Testing Strategy} assigns the same trained model to all the four agents. These agents share their explored map using a graph data structure to effectively collaborate and explore the entire environment together.

\section{Experiments and Results}
\label{sec:experiments-and-results}
In this section, we present the experimentation conducted to evaluate the effectiveness of our proposed approach. We describe the setup and metrics used to measure the performance of our method. 

\subsection{Experiments}
\label{sec:experiments:experiments}

To evaluate the effectiveness of our method, we conducted experiments on various environment sizes, including $\mathcal{G}{10}$, $\mathcal{G}{20}$, $\mathcal{G}{40}$, and $\mathcal{G}{60}$. We compared the performance of our approach, BAMAX, with traditional methods such as Depth-First Search (DFS) \cite{putri2011implementation} and Breadth-First Search (BFS) \cite{Rahim_2018}. To enable collaborative exploration, we extended the Distributed DFS algorithm \cite{MAKKI19967} and BFS algorithm \cite{beamer2012direction} to create Collaborative DFS and Collaborative BFS, respectively. These adaptations allow multiple agents to explore the grid map together. In our experiments, we generated 100 environments for each size, and in each scenario, we deployed 4 agents to explore the environment. Each agent started from a different random point, while ensuring a common starting point for all 4 agents across different methods, ensuring a fair comparison. We analyzed the results of the experiments based on following two key metrics.

\begin{enumerate}
  \item \textbf{Backtrack Count}  quantifies the frequency of backtracking performed by all agents, counting each transition from one point to another as a single backtrack step.
  
  \item \textbf{Simulation Steps} measures the count of timesteps taken collectively by all agents to explore the entire maze. Each simulation step represents a single step taken by all agents.
\end{enumerate}

\subsection{Results and Discussion}
\label{sec:experiments:results-and-discussion}

\begin{figure}[ht]
    \centering
    \subfloat[Grid $\mathcal{G}_{10}$\label{fig:performance:g10}] {{\includegraphics[width=10cm,height=3.5cm]{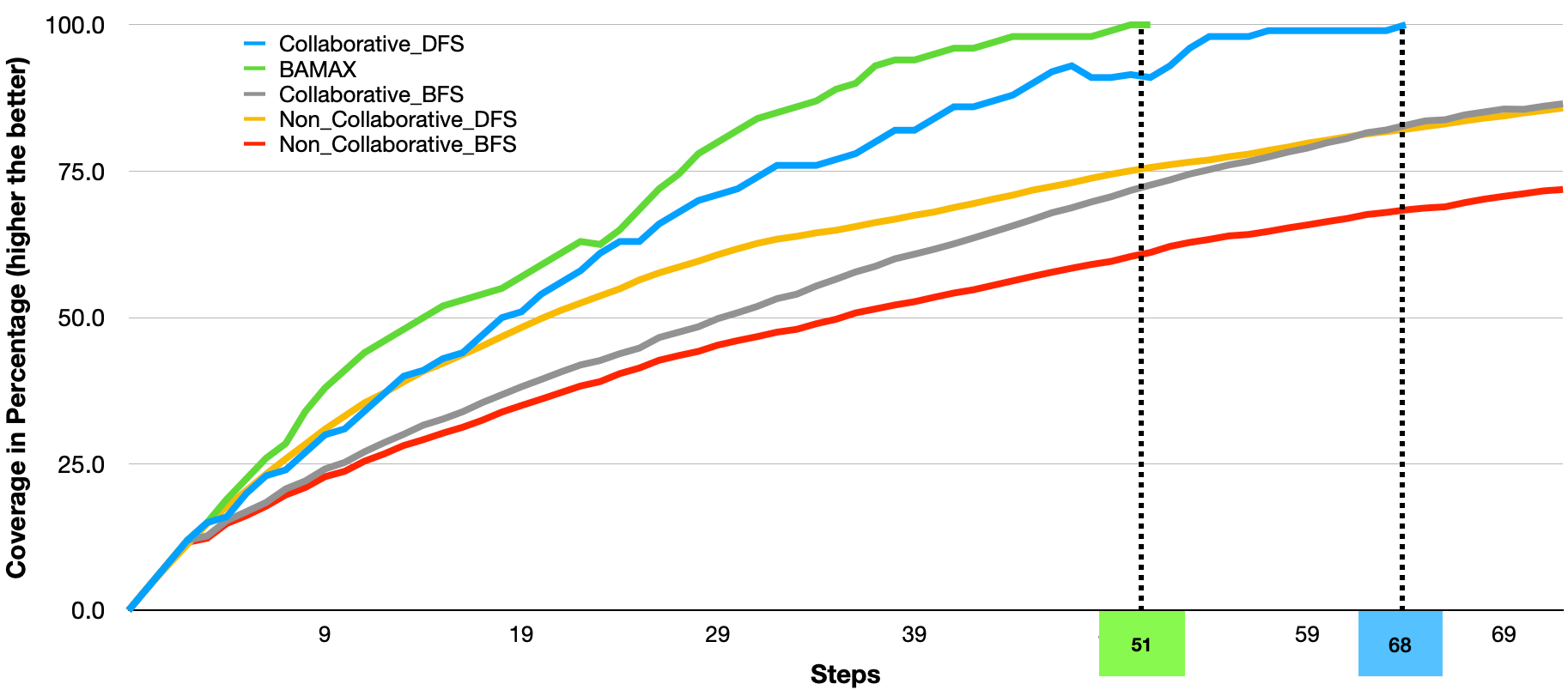} }}%
    \qquad
    \subfloat[Grid $\mathcal{G}_{60}$\label{fig:performance:g60}] {{\includegraphics[width=10cm,height=3.5cm]{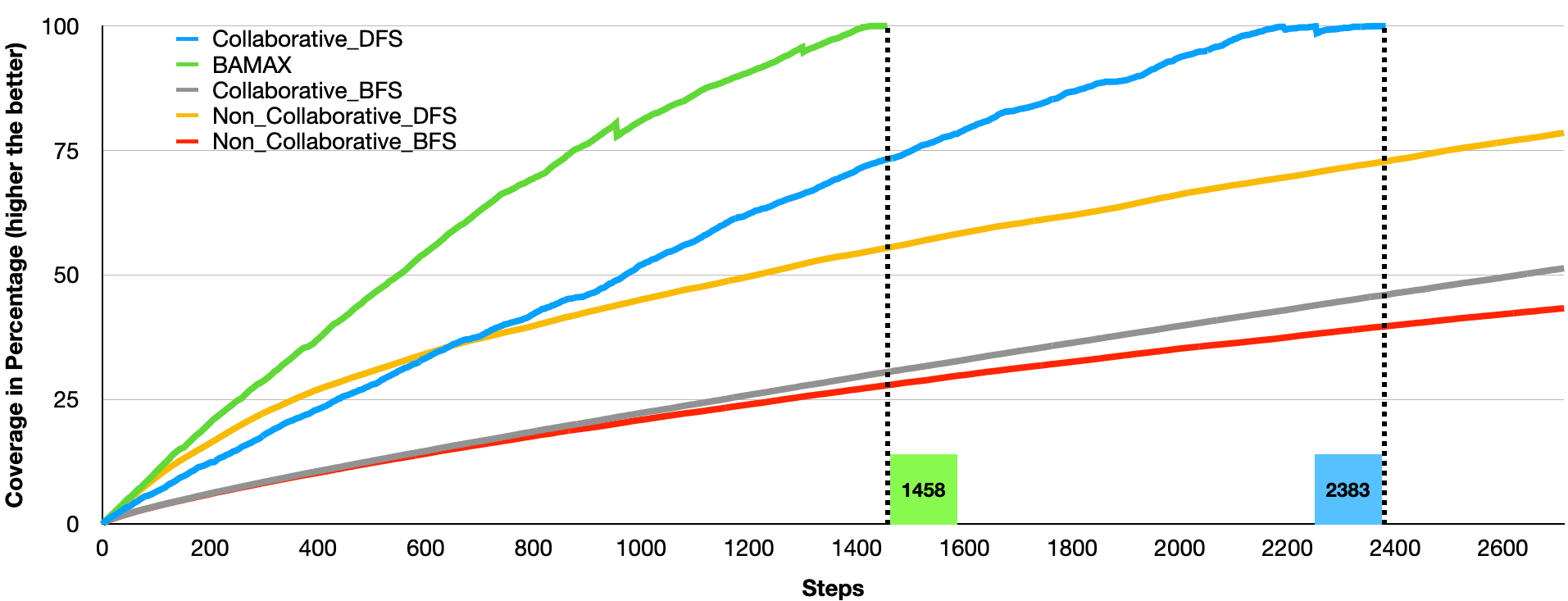} }}%
    \caption{Average performance of all methods on 100 hexagonal grids}%
    \label{fig:performance}%
\end{figure}

The performance of different methods across diverse environments is summarized in Table~\ref{tab:comparison}. The experimental results clearly demonstrate that our proposed method, BAMAX, outperforms traditional algorithms in terms of achieving faster 100 percent grid coverage with the minimum number of steps. The difference in the number of steps between BAMAX and the traditional algorithms is significant across various grid sizes. This could be attributed to the fact that algorithms like DFS and BFS are greedy approaches, which could prioritize certain paths without considering potentially better alternatives. In contrast, BAMAX understands the importance of effectively collaborating with other agents to optimize exploration. This significant difference arises from BAMAX's ability to intelligently select the most promising branches to explore, resulting in fewer steps required to cover the entire grid. 
\begin{table}[!ht]
\caption{Comparison of BAMAX with other methods}
\setlength{\tabcolsep}{2pt}
\renewcommand{\arraystretch}{1.5}
\label{tab:comparison}
\centering
\begin{tabular}{|l|cccc|cccc|}
\hline
\multicolumn{1}{|c|}{\multirow{2}{*}{\textbf{Method Name}}} &
  \multicolumn{4}{c|}{\textbf{Simulation Steps}} &
  \multicolumn{4}{c|}{\textbf{Backtrack Count}} \\ \cline{2-9} 
\multicolumn{1}{|c|}{} &
  \multicolumn{1}{c|}{$\mathcal{G}_{10}$} &
  \multicolumn{1}{c|}{$\mathcal{G}_{20}$} &
  \multicolumn{1}{c|}{$\mathcal{G}_{40}$} &
  $\mathcal{G}{60}$ &
  \multicolumn{1}{c|}{$\mathcal{G}_{10}$} &
  \multicolumn{1}{c|}{$\mathcal{G}_{20}$} &
  \multicolumn{1}{c|}{$\mathcal{G}_{40}$} &
  $\mathcal{G}{60}$ \\ \hline
DFS &
  \multicolumn{1}{c|}{149} &
  \multicolumn{1}{c|}{620} &
  \multicolumn{1}{c|}{2496} &
  \textgreater{}5900 &
  \multicolumn{1}{c|}{27} &
  \multicolumn{1}{c|}{117} &
  \multicolumn{1}{c|}{490} &
  803 \\ \hline
BFS &
  \multicolumn{1}{c|}{290} &
  \multicolumn{1}{c|}{\textgreater 1600} &
  \multicolumn{1}{c|}{\textgreater{}5900} &
  \textgreater 6100 &
  \multicolumn{1}{c|}{55} &
  \multicolumn{1}{c|}{228} &
  \multicolumn{1}{c|}{844} &
  1022 \\ \hline
Collaborative DFS &
  \multicolumn{1}{c|}{64} &
  \multicolumn{1}{c|}{276} &
  \multicolumn{1}{c|}{1061} &
  2383 &
  \multicolumn{1}{c|}{\textbf{8}} &
  \multicolumn{1}{c|}{36} &
  \multicolumn{1}{c|}{\textbf{150}} &
  345 \\ \hline
Collaborative BFS &
  \multicolumn{1}{c|}{182} &
  \multicolumn{1}{c|}{710} &
  \multicolumn{1}{c|}{2902} &
  \textgreater{}5900 &
  \multicolumn{1}{c|}{17} &
  \multicolumn{1}{c|}{78} &
  \multicolumn{1}{c|}{320} &
  712 \\ \hline
BAMAX &
  \multicolumn{1}{c|}{\textbf{51}} &
  \multicolumn{1}{c|}{\textbf{187}} &
  \multicolumn{1}{c|}{\textbf{644}} &
  \textbf{1458} &
  \multicolumn{1}{c|}{17} &
  \multicolumn{1}{c|}{\textbf{35}} &
  \multicolumn{1}{c|}{152} &
  \textbf{327} \\ \hline
\end{tabular}
\end{table}
 \\
Figure~\ref{fig:performance} shows that although all methods stay together at the start, BAMAX soon overtakes and stays in the lead. This is attributed to the fact that BAMAX agents are able to smartly select the exploration path which leads to lesser steps involved in backtracking. Even though collaborative DFS and collaborative BFS also have multiple collaborative agents, by the nature of their underlying exploration mechanism, collaborative DFS may face longer backtracking paths and collaborative BFS may carry out frequent backtracking from very early in the exploration leading to slower performance. This behavior can also be verified from backtracking counts in Table~\ref{tab:comparison}. Further, the difference in performance between BAMAX and other methods become even more significant as we increase the grid size. On $\mathcal{G}_{60}$ grid, BAMAX is able to achieve full grid coverage almost 38\% faster compared to next best method (collaborative DFS) as seen in figure~\ref{fig:performance:g60}.

\section{Conclusion}
\label{sec:conclusion}

This paper presents \textit{Backtrack Assisted Multi-Agent Exploration using Reinforcement Learning} (BAMAX) for collaborative exploration in hexagonal environments. Through extensive experimentation in environments of varying sizes, we compared BAMAX with traditional approaches. The results demonstrated BAMAX's capability to explore hexagonal grids consistent performance in exploration efficiency.
Further, the experimentation also showcases BAMAX's ability to translate its learning from training on a grid size of 10x10 to efficient explore hexagonal grids of different sizes.
This showcases good robustness and scalability across different environment sizes. Currently, BAMAX is able to extend its learning to grids of different sizes with only hexagonal cells. As part of future work, we would expand BAMAX's ability to handle grids with different sizes and also shapes.

\section{Reproducibility}
\label{sec:reproducibility}

After acceptance of this paper, we will open the source code, and provide documentation and instructions to reproduce the experiments discussed in this paper.


%
%
%
\bibliographystyle{splncs04}
\bibliography{document}

\end{document}